\pdfoutput=1
\documentclass[letterpaper, 10 pt, conference]{ieeeconf}  

\IEEEoverridecommandlockouts
\usepackage{times}
\usepackage{mathptmx}
\usepackage{amsmath,amssymb}
\usepackage{graphicx}
\usepackage{cite}
\usepackage{url}
\usepackage{xurl}
\usepackage{microtype}

\usepackage{pgfplots}
\pgfplotsset{compat=1.18}
\usepackage{tikz}
\usepackage{pgfplotstable}
\usepgfplotslibrary{fillbetween}
\usepackage{capt-of}

\let\labelindent\relax
\usepackage{enumitem}
\setlist[itemize]{leftmargin=*, itemsep=0pt, topsep=0.5pt, parsep=0pt, partopsep=0pt}
\setlist[enumerate]{leftmargin=*, itemsep=0pt, topsep=0.5pt, parsep=0pt, partopsep=0pt}

\title{\LARGE \bf
Needles in a Raystack: Ultra-Sparse LiDAR Occupancy Detection for Bat Tracks
}

\author{%
Nico Klar$^{1}$, Pankaj Rana$^{1}$, Nizam Gifary$^{1}$, Jakob Traub$^{1}$, Aamir Ahmad$^{2}$%
\thanks{$^{1}$Center for Solar Energy and Hydrogen Research (ZSW), Stuttgart, Germany.
        {\tt\small nico.klar@zsw-bw.de}}%
\thanks{$^{2}$University of Stuttgart, Institute of Flight Mechanics and Control (iFR), Flight Robotics and Perception Group (FRPG), Stuttgart, Germany.
        {\tt\small aamir.ahmad@ifr.uni-stuttgart.de}}%
}

\begin{document}
\maketitle
\thispagestyle{empty}
\pagestyle{empty}

\begin{abstract}

Monitoring flying animals is important for understanding and protecting biodiversity, but nocturnal species such as bats are difficult to observe in the field. Using LiDAR, bat movements at night result in ultra-sparse 3D spatio-temporal data in which standard reconstruction losses tend to predict only background and miss real flight paths. We study this problem as voxel-wise occupancy detection in sensor-centric LiDAR raystacks. A lightweight 3D U-Net is proposed that preserves temporal resolution, uses skip connections for spatial detail, and combines weighted binary cross-entropy with Dice loss to handle the strong class imbalance. 
In real LiDAR recordings of bats over open fields, cross-checked with acoustic monitoring, a reconstruction-based 3D convolutional autoencoder baseline fails to recover foreground trajectories. In contrast, the proposed U-Net recovers sparse foreground occupancy in diagnostic experiments and produces coherent occupancy patterns along bat flight trajectories, providing a practical basis for validation-scale experiments, later clustering of flight tracks, and future integration of bat activity information into biodiversity-aware turbine curtailment strategies.

\end{abstract}

\section{Introduction}
\label{sec:intro}

Flying animals such as bats are an important component of terrestrial biodiversity, yet their nocturnal flight activity remains difficult to monitor at relevant spatial and temporal scales. 
Onshore wind energy is a prominent conflict domain: a German field study of older turbines operating without curtailment estimated more than 70 bat fatalities per turbine over two months, while extrapolations for pre-regulation turbines suggest that annual fatalities in Germany could exceed 200{,}000 individuals if no effective mitigation measures are applied~\cite{voigt_wind_2022,pre-regulation}.
Reliable real-time detection is therefore needed to support biodiversity-aware turbine operation.

Current monitoring approaches provide complementary but incomplete capabilities.
Acoustic detectors are widely used, but their effective range is constrained by
ultrasonic attenuation, microphone sensitivity, call directionality, and species
identification ambiguity~\cite{voigt_limitations_2021}. Radar can cover larger
volumes around turbines, but bat detections are commonly validated using acoustic
recordings~\cite{krapivnitckaia_detection_2024, sato_detection_2025}. Thermal
camera systems have also been used for bat monitoring, often in multimodal setups
with ultrasonic sensors, but such workflows remain limited by range, observation
conditions, and review requirements~\cite{correia_bat_2013, cryan_behavior_2014}.
These limitations motivate sensing pipelines with stronger spatial localization and
automation capability.

LiDAR offers an active sensing modality for this setting: it provides 3D spatial
measurements, operates independently of ambient illumination, and can capture
fast-moving aerial activity near turbine-blade height~\cite{basics_lidar,
werber_drone-mounted_2023}. The field recordings used in this work were cross-checked with acoustic monitoring, so the retained trajectories can be attributed to bat activity.

\begin{figure}[t]
  \centering
  \includegraphics[width=0.9\linewidth]{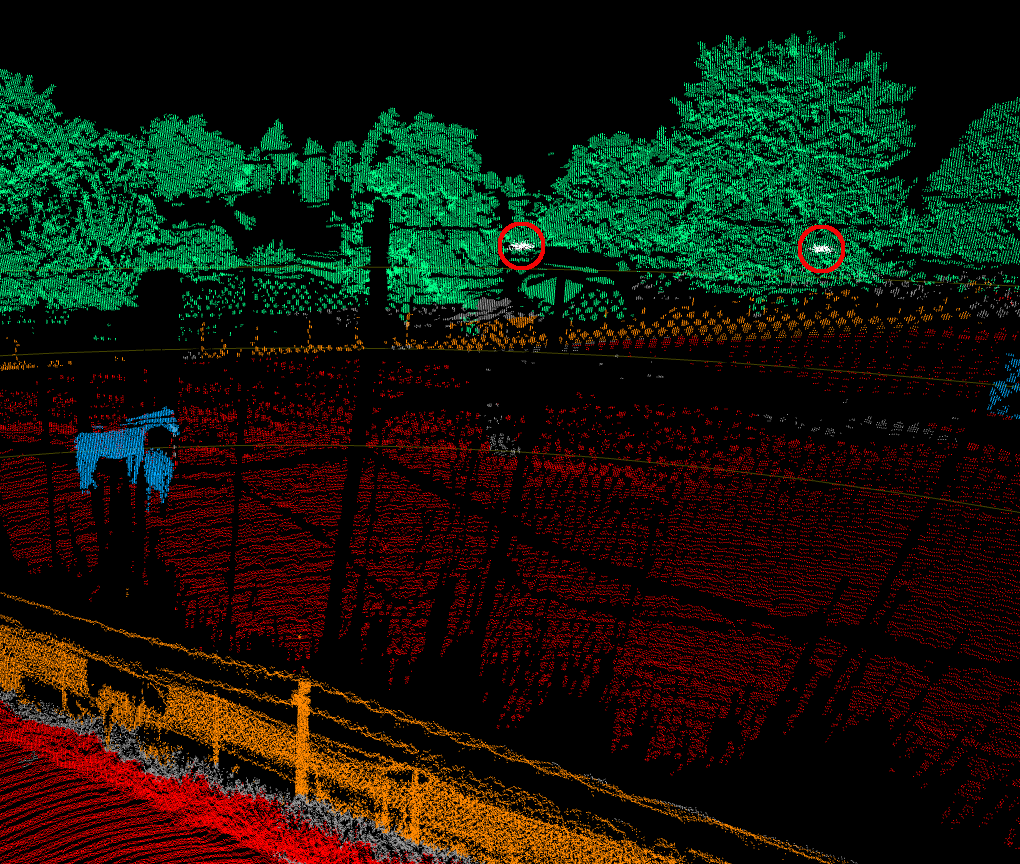}
    \caption{Representative outdoor LiDAR point cloud scene. The sparse white points highlighted by red circles indicate bat returns cross-checked with acoustic monitoring, illustrating how few foreground returns must be detected against a dense static background.}
  \label{fig:outdoor_lidar_scene}
\end{figure}

However, learning from LiDAR bat tracks introduces a
severe sparsity challenge. After preprocessing and background subtraction, bat
returns form ultra-sparse spatio-temporal grid sequences in which the occupied voxel
fraction is on the order of $10^{-5}$. Standard reconstruction objectives
therefore encourage all-background predictions, while aggressive downsampling can
remove the fine-grained localization information needed for single-pixel
trajectories. Recent LiDAR representation-learning methods show that occupancy
prediction is an effective objective for sparse point cloud data~\cite{boulch_also_2023,
min_occupancy-mae_2024}. Here, we study direct occupancy detection for bat-track
raystacks. We refer to this sensor-centric spatio-temporal grid representation as a
\emph{raystack}, because each frame stores sparse LiDAR returns in a fixed grid aligned
with the sensor view.

This work addresses \textbf{occupancy detection in ultra-sparse LiDAR raystacks} for
bat monitoring in open-field settings, with wind turbines considered as a key
application domain. Each track is represented as a 5D tensor over time and spatial
grid with per-return attributes (height, intensity, range) encoded as channels, and
detection is reformulated as per-voxel binary occupancy prediction. The main
contributions are:
\begin{itemize}
    \item A systematic ablation of failure modes using threshold sweeps, shift scans, and single-sample overfitting tests, showing that the autoencoder baseline fails because of architectural information loss rather than threshold miscalibration or spatial misalignment.
    \item A lightweight 3D U-Net with skip connections, no temporal pooling, and
    anisotropic $(1,3,3)$ kernels, paired with a combined weighted BCE and Dice loss
    to handle extreme class imbalance without collapsing to all-negative or
    all-positive predictions.
    \item Experimental evidence on real bat-track raystacks that sparse occupancy localization is learnable in diagnostic experiments, establishing a stable foundation for downstream track clustering and, in future work, for integrating LiDAR-based bat activity signals into automated turbine curtailment for biodiversity protection.
\end{itemize}

\section{Related Work}
\label{sec:related}

\subsection{Sensor-Based Monitoring of Nocturnal Aerial Fauna}

Monitoring flying animals at night has traditionally relied on acoustic detectors,
thermal cameras, radar systems, or combinations of these modalities. For bats,
acoustic monitoring is widely used because echolocation calls provide a direct and
often species-informative signal. However, acoustic range, call directionality,
species-specific call intensity, atmospheric attenuation, and environmental clutter
can introduce substantial detection biases. Recent reviews therefore emphasize
that no single fixed sensing modality is universally sufficient for bat monitoring
and that multimodal systems are often required to balance coverage, taxonomic
resolution, and robustness~\cite{he_structured_2025}.

Radar and thermal imaging have played an important role in nocturnal aeroecology.
Gauthreaux and Livingston combined a fixed vertical radar beam with a thermal
camera to estimate flight altitude and distinguish birds, bats, and insects from
track characteristics and altitude information~\cite{gauthreaux_monitoring_2006}.
More broadly, radar enables observations of birds, bats, and insects over spatial
scales that are difficult to access with local optical or acoustic sensors, but target
identity often remains ambiguous without complementary data~\cite{huppop_perspectives_2019}.
Camera-based deep learning systems have also been used to count and characterize
large bat emergences under deteriorating light conditions~\cite{koger_automated_2023},
and embedded vision platforms have recently been proposed for scalable terrestrial
biodiversity monitoring across multiple taxa~\cite{darras_eyes_2024}. These works
show the increasing role of automated sensing in biodiversity monitoring, while
also highlighting the limitations of purely image- or sound-based observations for
fine-scale three-dimensional flight analysis.

\subsection{LiDAR and Laser-Based Monitoring of Flying Animals}

LiDAR and laser-based sensing provide complementary capabilities for nocturnal
aerial monitoring because they can deliver spatially resolved observations without
visible illumination. Continuous-wave and Scheimpflug LiDAR systems have been
used to monitor flying insects with high temporal and spatial resolution, revealing
habitat-dependent abundance, size distributions, and diel activity patterns
~\cite{jansson_spatial_2023,tauc_wingbeat_2019}. Malmqvist et al. demonstrated
one of the closest ecological precedents to our setting by simultaneously monitoring
aerial insects, bats, and birds over a rice field using high-resolution Scheimpflug
LiDAR~\cite{malmqvist_batbirdbug_2018}. Their study showed that laser-based
systems can capture crepuscular and nocturnal activity patterns of interacting
aerial taxa in the field.

LiDAR has also been applied in bat-specific studies, although often for purposes
other than direct flight-path detection. Azmy et al. used terrestrial laser scanning
to count and spatially document roosting bats in caves under minimal light
conditions~\cite{azmy_counting_2012}. Hermans et al. combined acoustic tracking
with LiDAR-derived vegetation structure to study bat flight behaviour in
three-dimensional habitat context~\cite{hermans_combining_2023}. In wind-energy
research, Werber et al. used radar, LiDAR, and acoustic recorders to quantify bat
activity across altitude bands while testing a drone-mounted deterrent
~\cite{werber_drone-mounted_2023}. These studies motivate LiDAR as a useful
component of nocturnal bat monitoring, but they do not address learning-based
voxel-wise detection of bat returns in extremely sparse LiDAR sequences.

\subsection{Sparse Spatio-Temporal Occupancy Learning}

Extracting trajectories of small flying targets from sparse 3D observations remains
technically challenging. Nishisue et al. measured free-flying moth trajectories from
low-light infrared stereo data by constructing noise-reduced 3D point-cloud time
series and applying filtering steps to separate true target motion from artefacts
~\cite{nishisue_measuring_2024}. This illustrates a related problem: small nocturnal
flying animals produce sparse, noisy, and fragmented 3D observations that require
robust foreground extraction before ecological or behavioural analysis is possible.

In machine learning, sparse point-cloud and occupancy prediction methods often
use encoder--decoder or U-Net-like architectures to combine contextual aggregation
with localization-preserving skip connections~\cite{boulch_also_2023,min_occupancy-mae_2024}.
This is particularly relevant when the target signal occupies only a tiny fraction
of the input volume. In such settings, reconstruction losses can be dominated by
the empty background, while strong downsampling may remove the spatial evidence
needed to localize small foreground structures. For nocturnal bat LiDAR data, this
creates a joint ecological and technical problem: the sensor can provide valuable
3D information on bat activity, but the resulting raystack representation is so
sparse that standard reconstruction-based learning is poorly matched to the
detection objective. We therefore treat bat-track extraction as voxel-wise occupancy
detection and evaluate the model with threshold sweeps, alignment probes, and
single-sample overfitting tests to ensure that detected foreground voxels correspond
to recoverable flight trajectories rather than background-dominated predictions.

\section{Problem Formulation}
\label{sec:problem}

We consider learning-based representation and localization for \textbf{extremely sparse spatio-temporal track data}. Each sample is a fixed-length sequence of sensor-centric grid frames with shape $(T\times H\times W\times C)$ (in our setting $(T=50)$, $(H=W=1024)$, $(C=3)$). The signal is dominated by empty background: the vast majority of voxels are exactly zero, while only a tiny number of non-zero voxels appear per frame and form a moving trajectory across the full spatial extent. Fig.~\ref{fig:occupancy_and_unet_prediction} (left) illustrates this extreme sparsity using a zoomed-in view of a single binary occupancy target, where the foreground signal is reduced to a few isolated pixels in the original frame. This creates a learning problem that is strongly class-imbalanced at the voxel level and simultaneously sensitive to threshold selection, spatial misalignment, and architectural information loss.

\begin{figure}[t]
  \centering
  \includegraphics[width=0.90\linewidth]{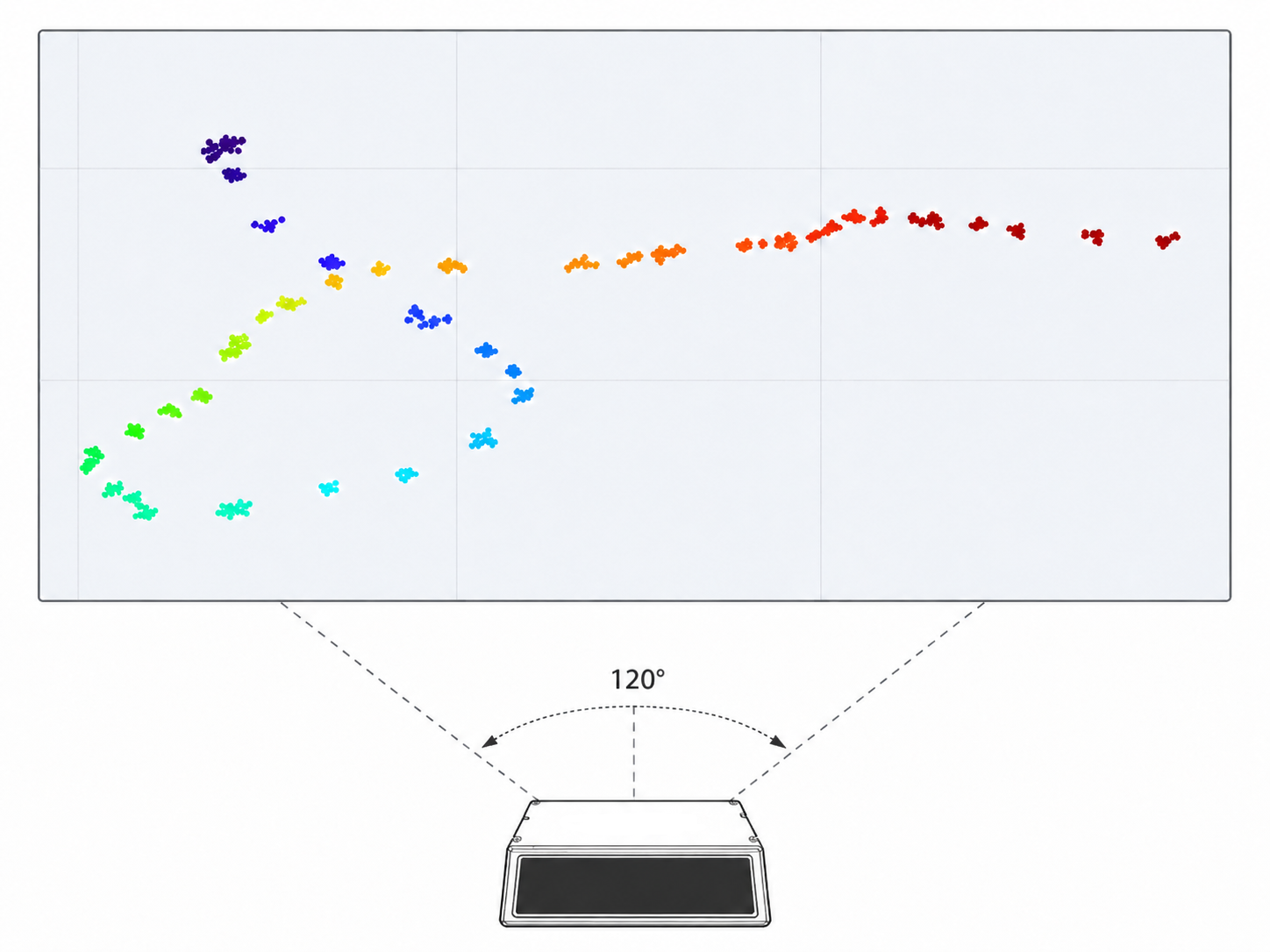}
  \caption{Bird's-eye-view projection of an example LiDAR track sample. Occupied voxels are projected onto the horizontal plane and color-coded by temporal progression, illustrating the ultra-sparse foreground trajectory used for voxel-wise occupancy detection.}
  \label{fig:bev_track_example}
\end{figure}

A direct reconstruction objective (e.g., MSE on intensity) is ill-suited in this regime. The model can reduce loss substantially by predicting near-zero everywhere, because the background dominates the error. However, for downstream analysis such as identifying consistent motion patterns or clustering track types, we require features that preserve whether and where a track exists across time. We therefore re-frame the learning target as \textbf{occupancy detection}: a binary mask indicating whether any non-zero value is present at voxel $(t,x,y)$, independent of which input channel was non-zero.

The central challenge is that positives are extremely rare.
In our data, the foreground occupancy is on the order of
$10^{-5}$ of all spatio-temporal grid cells; for $T = 50$ and
$H = W = 1024$, one sequence contains 52,428,800 voxel-wise
occupancy targets after aggregating over the input channels.
Such sparsity makes optimization brittle and encourages
degenerate solutions unless (i) the architecture preserves
localization information, (ii) the loss function reshapes gradients
under extreme imbalance, and (iii) the evaluation protocol goes
beyond a single thresholded metric.

Although the immediate learning target is binary voxel-wise occupancy, the model operates on complete spatio-temporal raystack sequences rather than on isolated frames or unordered point sets. The learned predictions therefore do not only indicate whether sparse LiDAR returns are present, but recover their temporally ordered occupancy structure across the sequence. This allows individual voxel predictions to form coherent flight trajectories and provides the basis for later track-level descriptors that capture temporal continuity, motion direction, fragmentation, and characteristic flight behaviours.

\section{Methodology}
\label{sec:method}

\subsection{Data Representation}
\label{sec:target}

Each sample is represented as a 5D tensor
$(\mathbf{X}\in\mathbb{R}^{T\times H\times W\times C})$. Values are effectively
near-binary in practice: most entries are exactly zero, while non-zero entries
occur only at a few spatial locations per frame. Prior to training, we apply
per-channel normalization only over non-zero entries to avoid shrinking the
sparse signal by statistics dominated by the empty background.

We define the binary occupancy mask
$(\mathbf{Y}\in\{0,1\}^{T\times H\times W\times 1})$ as
\begin{equation}
\label{eq:occ_target}
Y_{txy} = \mathbf{1}\Big[\exists c \in \{1,\dots,C\} : X_{txyc} \neq 0\Big].
\end{equation}
This target discards intensity and channel identity and focuses the learning
problem on spatio-temporal localization of the track.

Because $\mathbf{Y}$ can contain single-pixel positives, supervision can be
overly brittle, especially early in training. We therefore optionally apply
per-frame morphological dilation to enlarge positives:
\begin{itemize}
\item Dilation radius $(r=1)$ corresponds to a $(3\times 3)$ neighborhood.
\item Dilation radius $(r=2)$ corresponds to a $(5\times 5)$ neighborhood.
\end{itemize}
This increases the effective positive rate and yields less sparse gradients
without changing the underlying track topology.

\subsection{Baseline Autoencoder}
\label{sec:baseline}

Our initial baseline model is a reconstruction-based 3D convolutional
autoencoder that maps $(\mathbf{X})$ to a compact latent vector
$(\mathbf{h}\in\mathbb{R}^{d})$ and reconstructs a dense output at the original
spatio-temporal resolution. The approach is related in spirit to masked
occupancy autoencoding for LiDAR representation learning
~\cite{min_occupancy-mae_2024} and occupancy-based self-supervision
~\cite{boulch_also_2023}, but is used here as a direct baseline for
ultra-sparse bat-track raystacks rather than as a pretraining objective for
large-scale automotive LiDAR perception.

The encoder uses repeated Conv3D + Pool blocks followed by a Dense bottleneck,
while the decoder mirrors this structure with transposed convolutions. Key
design choices are:
\begin{itemize}
\item 3D convolutions with kernel $(3,3,3)$.
\item Aggressive pooling in space and early pooling in time to reduce $(T)$.
\item Flattening followed by a Dense bottleneck for compact latent codes.
\end{itemize}
This reconstruction-oriented design is a reasonable baseline for dense
spatio-temporal data, but it is poorly aligned with tiny-object localization
under extreme foreground--background imbalance.

Given the sparsity, MSE on intensities is inappropriate. We therefore trained
the baseline with occupancy-oriented losses derived from the model output. In
the following, $i$ indexes spatio-temporal grid cells, $y_i$ denotes the binary
occupancy label, and $\ell_i$ denotes the predicted occupancy logit.

\begin{enumerate}
\item \textbf{Weighted binary cross-entropy (BCE) with logits.}
BCE denotes binary cross-entropy, equivalently the Bernoulli negative
log-likelihood for binary labels. We use a positive-class weight to compensate
for the rarity of occupied voxels:
\begin{equation}
\label{eq:wbce}
\begin{aligned}
\mathcal{L}_{\mathrm{wBCE}}
= -\frac{1}{N}\sum_i
\big[&
p\,y_i \log \sigma(\ell_i) \\
&+ (1-y_i)\log(1-\sigma(\ell_i))
\big],
\end{aligned}
\end{equation}
where $\sigma(\cdot)$ is the sigmoid function and
$p=N_{\mathrm{neg}}/N_{\mathrm{pos}}$ is computed from the numbers of
background and occupied voxels. To prevent numerical issues under extreme
imbalance, $p$ is clipped to a maximum value.

\item \textbf{Focal loss with logits.}
As an alternative imbalance-aware objective, we also tested focal loss:
\begin{equation}
\label{eq:focal}
\mathcal{L}_{\mathrm{focal}}
= -\frac{1}{N}\sum_i \alpha_{t,i}
(1-p_{t,i})^\gamma \log(p_{t,i}),
\end{equation}
with
\begin{equation}
p_{t,i}
=
y_i\sigma(\ell_i)
+
(1-y_i)(1-\sigma(\ell_i)),
\end{equation}
and
\begin{equation}
\alpha_{t,i}
=
\alpha y_i + (1-\alpha)(1-y_i).
\end{equation}
Here, $p_{t,i}$ is the probability assigned to the ground-truth class. We
initially used a high positive-class weighting factor, e.g. $\alpha=0.99$, to
emphasize rare occupied voxels.
\end{enumerate}

\subsection{Proposed 3D U-Net}
\label{sec:unet}

To restore localization ability, we replace the Dense bottleneck autoencoder
with a fully convolutional, skip-connected encoder--decoder in the style of a
U-Net. The model is designed to preserve high-resolution spatial information
while still aggregating local context.

\textbf{Design goals}:
\begin{enumerate}
\item Preserve localization of single-pixel structures.
\item Avoid temporal pooling to maintain time alignment.
\item Control compute using anisotropic kernels.
\end{enumerate}

\textbf{Key components}:
\begin{itemize}
\item Encoder depth $(D)$ with repeated blocks of two Conv3D layers.
\item Pooling only in space: $(1,2,2)$, leaving $(T)$ unchanged.
\item Decoder upsampling with Conv3DTranspose $(1,2,2)$ and concatenation of
corresponding encoder activations via skip connections.
\item One-channel occupancy logits as output, computed in float32.
\item Anisotropic kernels $(1,3,3)$ to emphasize spatial context while keeping
compute feasible at $(1024\times 1024)$ resolution.
\end{itemize}

\begin{figure*}[t]
  \centering
  \includegraphics[width=0.95\textwidth]{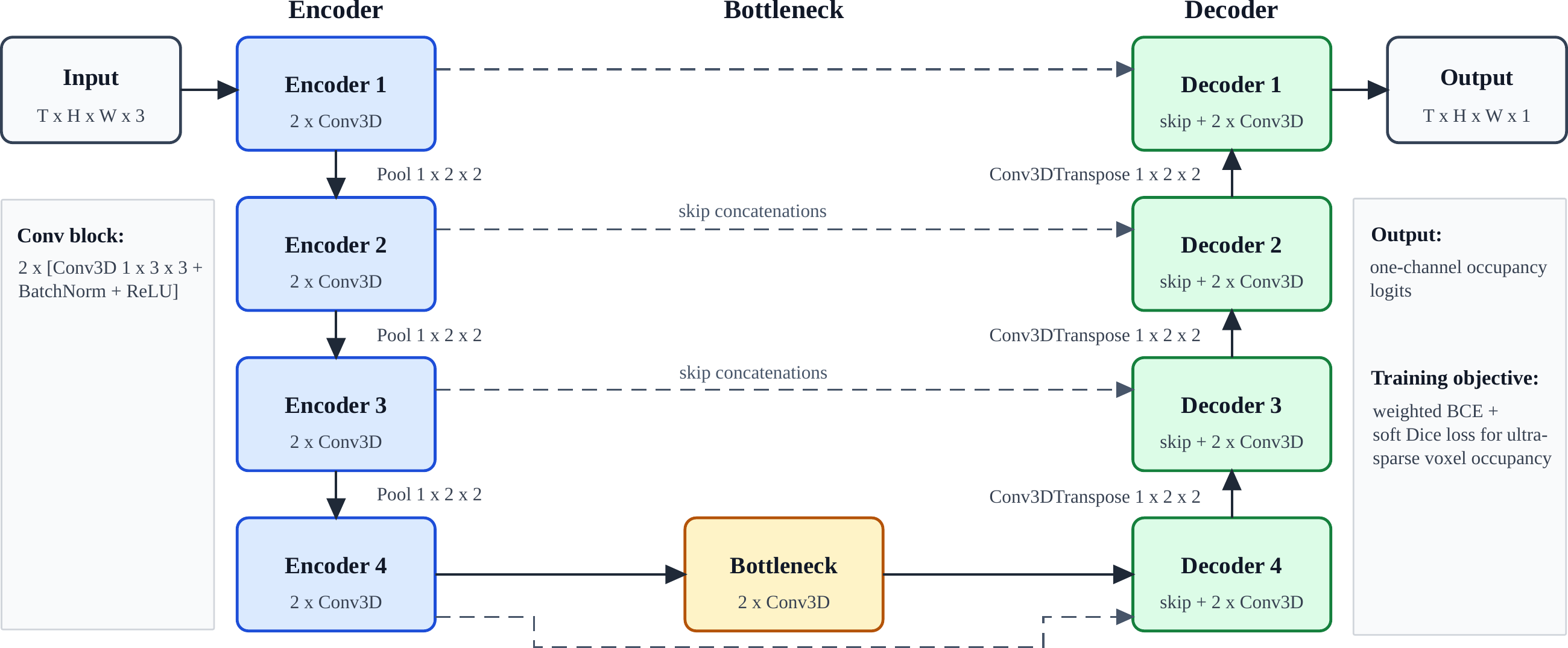}
  \caption{Architecture of the proposed 3D U-Net for voxel-wise occupancy prediction in ultra-sparse LiDAR raystacks. The U-shaped encoder--decoder preserves the temporal dimension, pools and restores spatial resolution, and uses skip concatenations to retain localization information.}
  \label{fig:unet_architecture}
\end{figure*}

Compared to the baseline, the key inductive-bias shift is from global compression
and reconstruction to dense per-voxel detection with feature reuse. To avoid both
all-negative and all-positive collapse, we combine the weighted BCE from
Eq.~\eqref{eq:wbce} with Dice loss. Dice loss is computed on sigmoid probabilities
$\hat{y}_i=\sigma(\ell_i)$:
\begin{equation}
\label{eq:dice}
\mathcal{L}_{\mathrm{Dice}}
=
1 -
\frac{
2\sum_i \hat{y}_i y_i + \epsilon
}{
\sum_i \hat{y}_i + \sum_i y_i + \epsilon
},
\end{equation}
where $\epsilon$ is a small constant for numerical stability. The final objective is
\begin{equation}
\label{eq:combo_loss}
\mathcal{L}
=
\mathcal{L}_{\mathrm{wBCE}}
+
\mathcal{L}_{\mathrm{Dice}}.
\end{equation}
Conceptually, weighted BCE provides per-voxel gradients with increased weight on
rare occupied voxels, while Dice loss directly optimizes foreground overlap and
helps suppress trivial background- or foreground-dominated solutions.

\subsection{Evaluation and Failure-Mode Ablation}
\label{sec:metrics}

In ultra-sparse detection, a single threshold can hide failure modes. We therefore
evaluate occupancy predictions at multiple thresholds
$(\tau\in\{0.5,0.2,0.1,0.05,0.02\})$ using pixel-level precision, recall, F1, and
predicted positive rate. We further use mean predicted probability on
ground-truth positives versus negatives, Top-$K$ hit-rate, shift scans in
$(\Delta t,\Delta x,\Delta y)$, and single-sample overfitting tests. Together,
these analyses separate threshold calibration, spatial misalignment, and
localization failure as possible causes of poor foreground recovery.

\section{Experiments and Results}
\label{sec:results}

\subsection{Experimental Setup}
\label{sec:setup}

Track sequences are loaded from compressed NumPy arrays and stacked into
$(N,T,H,W,C)$. Per-channel normalization is applied on non-zero entries only.
The data are split at the trajectory level to avoid leakage between training
and validation samples. We use a 75\%/25\%
train--validation split to preserve the distribution of sparse
foreground trajectories across both subsets.

\begin{itemize}
\item Optimizer: Adam.
\item Learning rate: $10^{-4}$.
\item Due to the large spatial resolution of the raystack inputs, training is
performed with batch size 1. Validation is also performed with batch size 1.
\item Early stopping and checkpointing are based on a detection-oriented
validation metric, such as validation F1 at $\tau=0.5$, rather than solely on
$\text{val\_loss}$.
\end{itemize}

\subsection{Autoencoder-Based Occupancy Reconstruction}
\label{sec:baseline_results}

During baseline training, the loss decreased, which superficially suggested progress. However, evaluation consistently showed:
\begin{itemize}
\item $(\text{TP}=0)$ across thresholds.
\item Precision/Recall/F1 remained 0.
\item Mean predicted probability on ground-truth positives was not greater than on ground-truth negatives.
\item Top-$K$ hit-rate was 0.
\end{itemize}
This indicates that the baseline reduced loss by modeling the dominant background distribution rather than learning track locations.

A shift scan over small $(\Delta t,\Delta x,\Delta y)$ offsets did not recover any true positives, ruling out a pure alignment error. More importantly, the single-sample overfitting test failed, confirming that the architecture was poorly suited to tiny-object localization. We tested a wide range of thresholds. While predicted-positive counts changed dramatically with $(\tau)$, the true-positive count remained 0 in the baseline, indicating that the model's scoring function was not correlated with the occupancy target.

We observed two failure modes across losses:
\begin{itemize}
\item \textbf{All-negative collapse}: extremely low predicted-positive rate at standard thresholds (e.g., $(\tau=0.5)$).
\item \textbf{All-positive collapse}: saturated predicted-positive rate near 1.0 for low thresholds when using a high focal-loss $(\alpha)$.
\end{itemize}
These collapse modes motivated the combined BCE--Dice objective used in the final model.

\subsection{U-Net-Based Occupancy Detection}
\label{sec:proposed_results}

We first tested whether the ultra-sparse occupancy target is learnable at all by
overfitting a single repeated training sample. This removes generalization from
the analysis and directly probes model capacity and inductive bias. Unlike the
baseline, the proposed model recovered the foreground trajectory and clearly
separated occupied from empty voxels.

Representative results on the training sample:
\begin{itemize}
\item At $(\tau=0.5)$: Precision 0.833, Recall 0.844, F1 0.838 (TP 1671, FP 336, FN 309). 
\item Probability separation: mean $(p)$ on ground-truth positives $(\approx 0.765)$ vs mean $(p)$ on ground-truth negatives $(\approx 1.1\times 10^{-5})$. 
\item Top-10,000 hit-rate $(\approx 0.198)$, indicating strong enrichment of ground-truth positives among the top-ranked voxels.
\end{itemize}
These results show that the U-Net architecture and the combined loss can recover sparse foreground occupancy in a controlled diagnostic setting when generalization effects are removed.

The threshold sweep shows the expected precision--recall trade-off:
\begin{itemize}
\item $(\tau=0.5)$: high precision with strong recall.
\item Lower $(\tau)$ values increase recall further but admit more false positives.
\end{itemize}
This allows the operating threshold to be selected according to the application requirement, e.g., conservative detection or increased recall.

\begin{center}
  \begin{minipage}{\linewidth}
    \centering
    \begin{minipage}[t]{0.49\linewidth}
      \centering
      \includegraphics[
        width=\linewidth,
        height=0.17\textheight,
        keepaspectratio
      ]{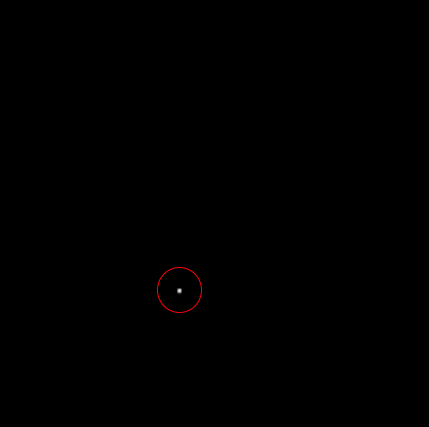}
    \end{minipage}
    \hfill
    \begin{minipage}[t]{0.49\linewidth}
      \centering
      \includegraphics[
        width=\linewidth,
        height=0.17\textheight,
        keepaspectratio
      ]{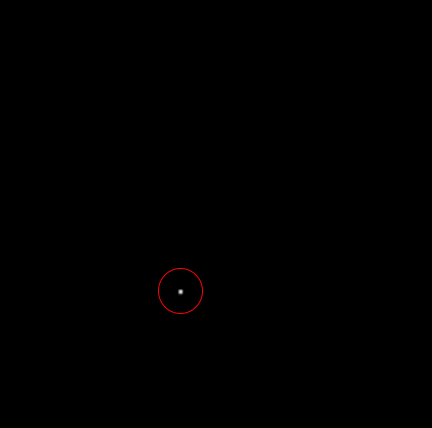}
    \end{minipage}
    \captionof{figure}{Dilated ground-truth occupancy target and qualitative U-Net occupancy prediction for a sparse bat trajectory sample. Both panels show a zoomed crop of the original occupancy frame. The left panel shows the dilated target for a bat detected at approximately 35 m from the LiDAR, while the right panel shows the corresponding model prediction.}
    \label{fig:occupancy_and_unet_prediction}
  \end{minipage}
\end{center}

Beyond scalar metrics, qualitative inspection of predicted masks shows that the model concentrates probability mass tightly around true track locations, rather than distributing probability uniformly over the frame. This matches the quantitative separation of mean probabilities and supports that the model captures localization cues rather than only background statistics.

\subsection{Discussion}
\label{sec:why}

The improvement can be attributed to three design choices. First, skip connections reintroduce high-resolution encoder features in the decoder, so the model does not need to recover spatial detail from a compressed latent vector. Second, pooling only in $(H,W)$ preserves temporal indices and reduces the risk of time shifts for moving points. Third, the combined loss addresses the extreme foreground--background imbalance: weighted BCE increases the contribution of rare occupied voxels, while Dice loss directly optimizes foreground overlap. These choices target the main limitations identified in the baseline: spatial information loss, temporal downsampling effects, and foreground suppression.

\subsection{Future Work}
\label{sec:future_work}

Future work will extend the evaluation to larger and more diverse field recordings, including independent recording sessions and additional environmental conditions. We will refine operating-threshold selection using validation F1 or precision--recall analysis and study the effect of target dilation on recall, precision, and localization sharpness. Further ablations should compare anisotropic $(1,3,3)$ kernels with $(3,3,3)$ kernels and vary model capacity to trade accuracy against runtime. In addition to voxel-level metrics, future evaluations should include track-level criteria such as trajectory continuity, fragmentation, false-positive track formation, spatial deviation, and distance-dependent detection performance.

\section{Conclusion}
\label{sec:conclusion}

We investigated learning on ultra-sparse spatio-temporal track data and found that a 3D convolutional autoencoder with aggressive downsampling and a Dense bottleneck fails to localize single-pixel trajectories, even under single-sample overfitting. Failure-mode ablations indicate that the root cause is architectural information loss rather than threshold calibration or spatial misalignment.

A lightweight 3D U-Net with skip connections, no temporal pooling, anisotropic kernels, and a combined weighted BCE--Dice loss successfully recovers sparse occupancy in controlled diagnostic experiments under extreme imbalance. The proposed approach establishes a stable foundation for validation-scale generalization experiments and subsequent track-level clustering pipelines.

\section*{Acknowledgment}
This work was funded by the German Federal Ministry for Economic Affairs and Climate Action based on a decision of the German Bundestag under grant 03EE2047A.

\bibliographystyle{IEEEtran}
\bibliography{literature}

\end{document}